\documentclass[10pt,twocolumn,letterpaper]{article}

\usepackage{wacv}
\usepackage{times}
\usepackage{epsfig}
\usepackage{graphicx}
\usepackage{amsmath}
\usepackage{amssymb}
\usepackage{booktabs}
\usepackage{subcaption}
\usepackage{multirow}
\usepackage{float}
\usepackage{booktabs}
\usepackage{enumitem}

\usepackage{pifont}
\newcommand{\cmark}{\ding{51}}%
\newcommand{\xmark}{\ding{55}}%

\usepackage[accsupp]{axessibility}

\def\wacvPaperID{236} 

\wacvalgorithmstrack   

\ifwacvfinal
\usepackage[breaklinks=true,bookmarks=false]{hyperref}
\else
\usepackage[pagebackref=true,breaklinks=true,colorlinks,bookmarks=false]{hyperref}
\fi

\wacvfinalcopy

\begin{document}
\title{ARUBA: An Architecture-Agnostic Balanced Loss for Aerial Object Detection
}

\author{Rebbapragada V C Sairam~~~~~~ Monish Keswani~~~~~~ Uttaran Sinha \\Nishit Shah~~~~~~ Vineeth N Balasubramanian\\
Indian Institute of Technology Hyderabad \\
{\tt\small \{ai20resch13001, monish.keswani, cs17mtech11003, cs18mtech11020, ~vineethnb\}@iith.ac.in} \\
\vspace{-12pt}
}

\maketitle
\thispagestyle{empty}

\begin{abstract}
\vspace{-1.5mm}
   Deep neural networks tend to reciprocate the bias of their training dataset. In object detection, the bias exists in the form of various imbalances such as class, background-foreground, and object size. In this paper, we denote size of an object as the number of pixels it covers in an image and size imbalance as the over-representation of certain sizes of objects in a dataset. We aim to address the problem of size imbalance in drone-based aerial image datasets.  Existing methods for solving size imbalance are based on architectural changes that utilize multiple scales of images or feature maps for detecting objects of different sizes. We, on the other hand, propose a novel \textbf{AR}chitect\textbf{U}re-agnostic \textbf{BA}lanced Loss (\textbf{ARUBA}) that can be applied as a plugin on top of any object detection model. It follows a neighborhood-driven approach inspired by the ordinality of object size.  We evaluate the effectiveness of our approach through comprehensive experiments on aerial datasets such as HRSC2016, DOTAv1.0, DOTAv1.5 and VisDrone and obtain consistent improvement in performance. 
\end{abstract}

\vspace{-4mm}
\section{Introduction}
\vspace{-1mm}
In recent years, drones have shown immense potential in numerous disciplines. In military warfare, they can be used as target decoys for combat missions. In agriculture, drones provide farmers with real-time data to make informed harvesting decisions. For search-and-rescue, they can reach places where humans cannot. Alternatively, they are also used in fire-fighting, delivery of essentials and aerial photography. This increasing demand for drones in various domains has recently encouraged the computer vision community to work extensively on vision from drones \cite{cazzato2020survey}. 

Deep neural networks have led computer vision research and development for a decade now on multiple challenging problems such as semantic segmentation, object detection/tracking, as well as image classification. In object detection, methods like FasterRCNN \cite{Ren2015FasterRT}, YOLO \cite{Redmon2016YouOL}, RetinaNet \cite{Lin2017FocalLF} and its variants have achieved decent performance on many challenging datasets. 
With increased interest and creation of datasets in drone-based imagery, aerial object detection \cite{Xia2018DOTAAL,Du2019VisDroneDET2019TV} has gained a lot of interest from the research community. Although the aforementioned methods exhibit exceptional performance on popular general object detection datasets such as MSCOCO \cite{lin2014microsoft}, 
aerial-object datasets  \cite{Xia2018DOTAAL,Du2019VisDroneDET2019TV} pose more challenges, even to state-of-the-art object detection models. 

\begin{figure}[t]
    \centering
    \includegraphics[width=\linewidth]{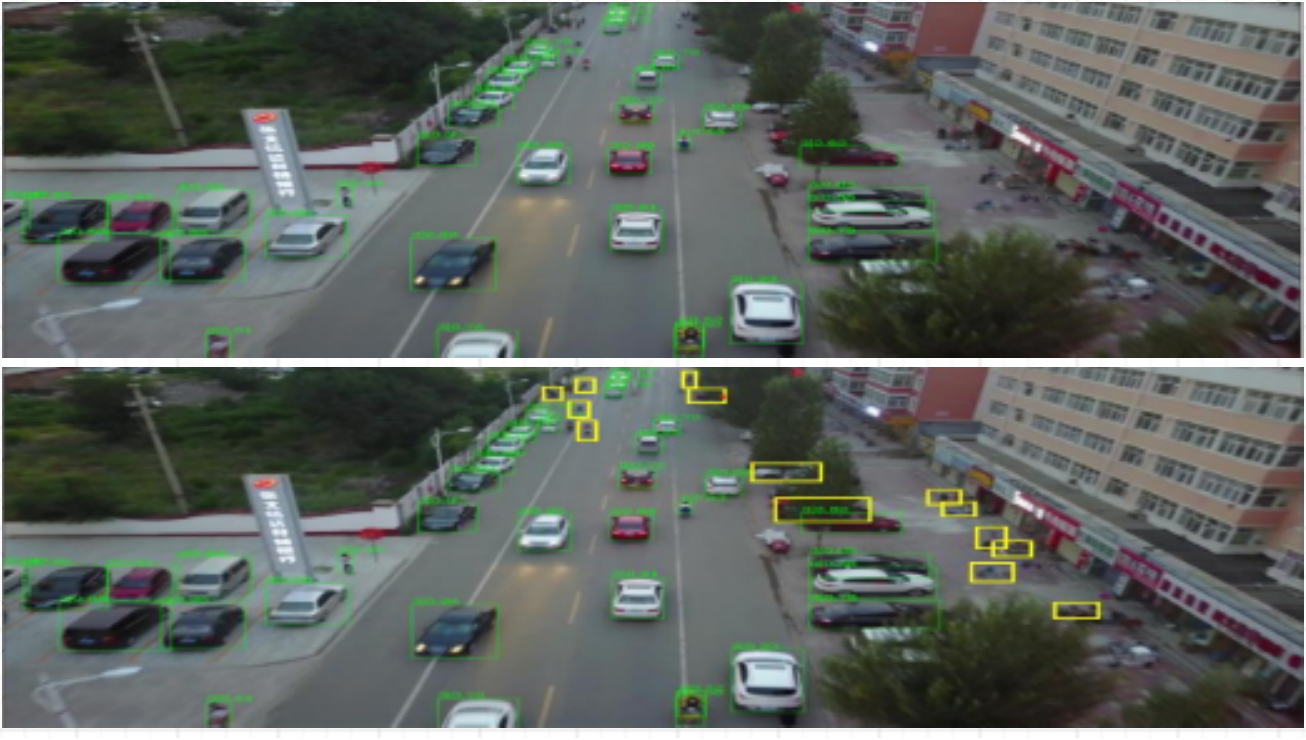}
    \vspace{-5mm}
    \caption{Predictions on an image from VisDrone dataset \cite{Du2019VisDroneDET2019TV} with Focal loss \cite{lin2017focal} vs Ours. \textbf{Top:} Focal loss fails to detect many objects. \textbf{Bottom:} Ours is able to recognize additional objects, including small ones, because of our \textbf{AR}chitect\textbf{U}re-agnostic \textbf{BA}lanced (\textbf{ARUBA}) loss. Yellow boxes indicate objects additionally detected.}
    \label{demo}
    \vspace{-5mm}
\end{figure}

\begin{figure}[ht]
\centering
\begin{subfigure}{.22\textwidth}
  \centering
  \includegraphics[width=\linewidth]{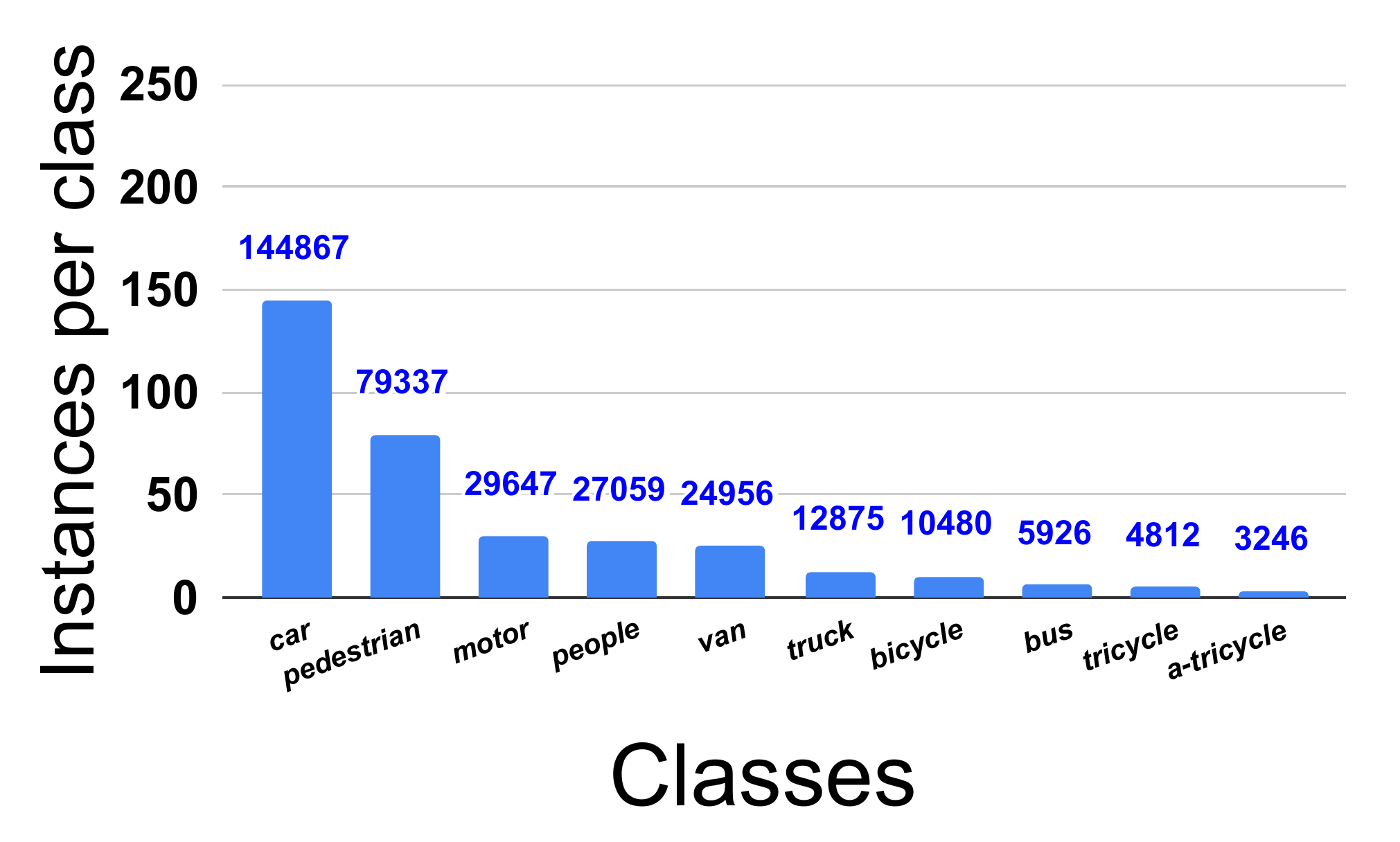}  
  \caption{\centering Class Imbalance}
  \label{fig:vis-class-imb}
\end{subfigure}
\hfill
\begin{subfigure}{.22\textwidth}
  \centering
  \includegraphics[width=\linewidth]{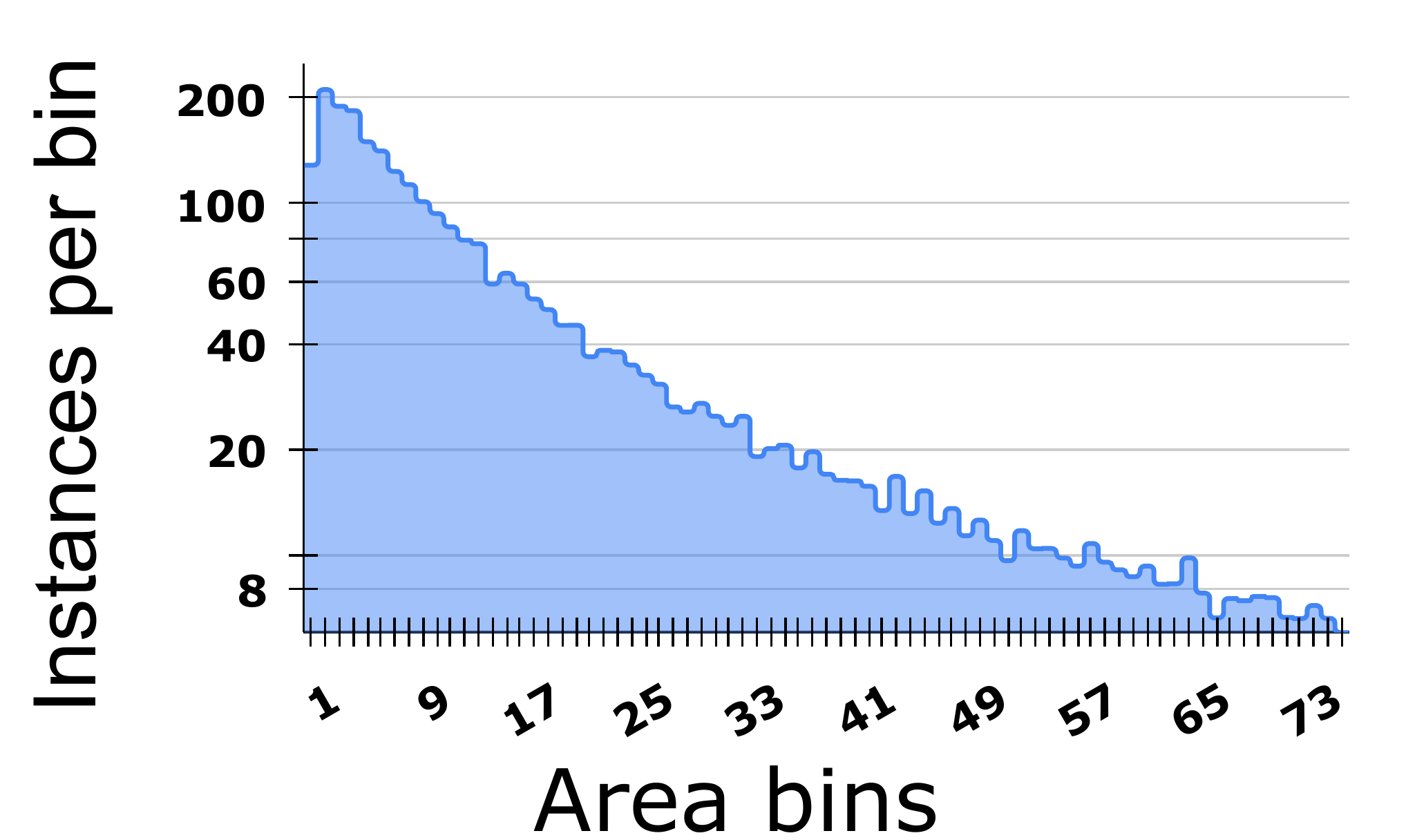}  
  \caption{\centering Size Imbalance}
  \label{fig:vis-size-imb}
\end{subfigure}
\vspace{-2mm}
\caption{Highly skewed class and size distributions in VisDrone dataset}
\vspace{-5.5mm}
\label{fig:cls-size-imb-comp}
\end{figure}

\begin{figure}[ht]
\centering
\begin{subfigure}{.22\textwidth}
  \centering
  \includegraphics[width=\linewidth]{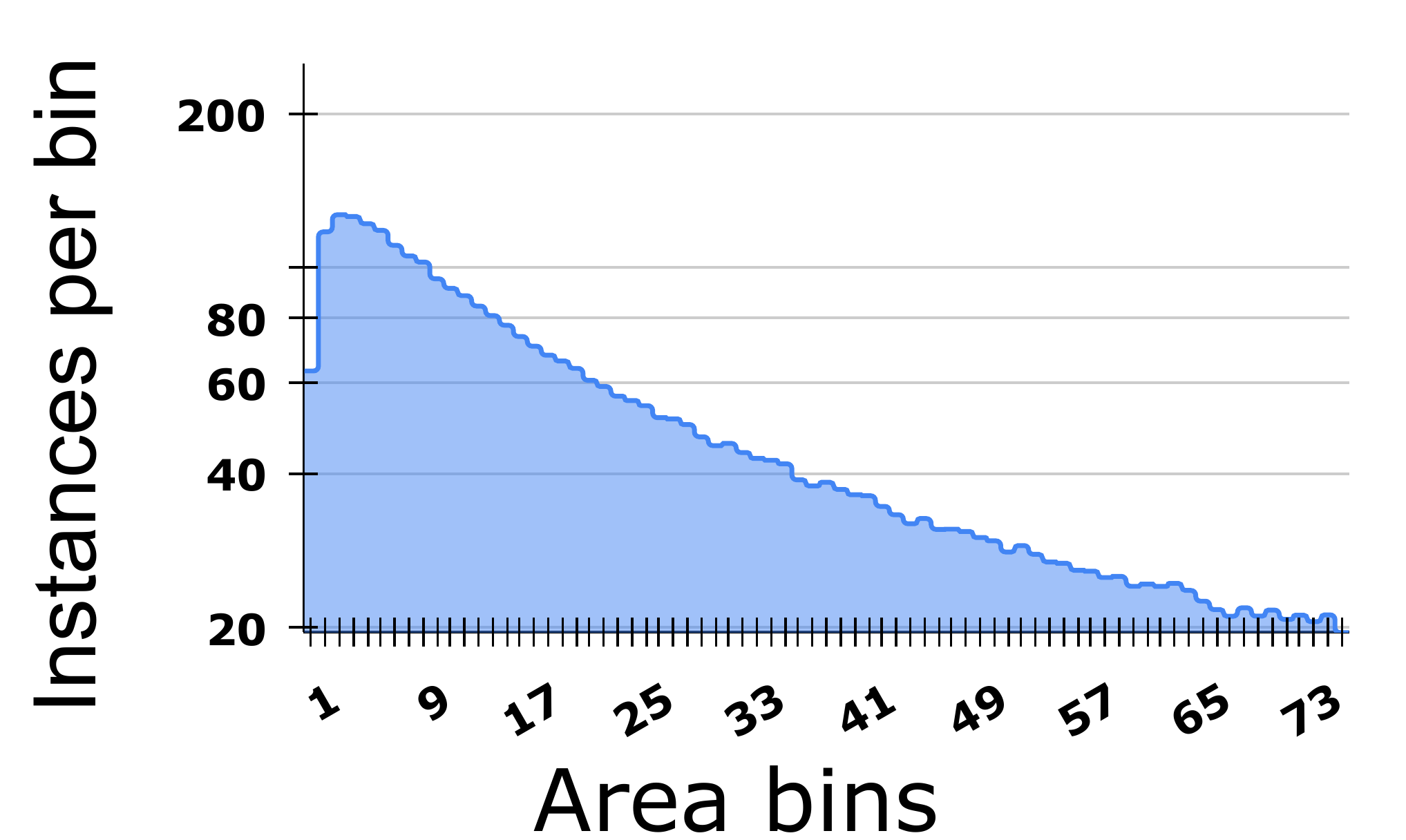}  
  \caption{\centering Size Imbalance-COCO}
  \label{fig:coco-size-imb}
\end{subfigure}
\hfill
\begin{subfigure}{.22\textwidth}
  \centering
  \includegraphics[width=\linewidth]{charts/VisDrone_area_imb.pdf}  
  \caption{\centering Size Imbalance-VisDrone}
  \label{fig:vis-size-imb-1}
\end{subfigure}
\vspace{-2mm}
\caption{Comparison of size imbalance severity between general and drone-based aerial object datasets. 
Note that $y$-axis is log of frequency, hence the effect is exponential in terms of occurrence.}
\label{fig:size-imb-comp}
\vspace{-5.5mm}
\end{figure}

High variation in scale and orientation of objects in aerial datasets, especially from drone images, make detecting these objects quite challenging. Specialized methods \cite{han2021redet,yi2021oriented} have been proposed to capture the oriented bounding boxes efficiently. An added difficulty in aerial datasets \cite{Xia2018DOTAAL,Du2019VisDroneDET2019TV,liu2017high} is that they are highly skewed in their object size distribution in addition to the class distribution (as shown in Figure \ref{fig:cls-size-imb-comp}. Note that in Figure \ref{fig:vis-size-imb}, x-axis shows the object area bins where the size of the objects increases from left to right and y-axis shows the number of object instances per an area bin). We also observe that size imbalance is severe in aerial object datasets when compared to more general-purpose object detection datasets (refer Figure \ref{fig:size-imb-comp}), which motivates us to address this imbalance problem of drone-based aerial datasets in this work.

Size imbalance is a common problem in object detection datasets, and many methods have been proposed to mitigate this issue, as summarized in \cite{oksuz2020imbalance}. Existing methods \cite{Lin2017FeaturePN,liu2016ssd,singh2018analysis} have largely proposed architectural modifications to enhance the model's ability to view objects at different scales. However, such multi-scale approaches arise from careful engineering of architectures to suit a specific domain or setting. In this work, we propose to address the size imbalance problem from an architecture-agnostic balanced loss perspective. One could also view our approach as a long-tailed perspective to a size balance problem, unlike the class imbalance setting that is typically studied in long-tailed detection/recognition problems.
size imbalance in aerial datasets. In contrary to the existing methods, we propose an architecture-independent approach which can be applied as a plugin on top of any object detection method.

Long-tailed object detection methods typically focus on datasets with skewed class distribution, to improve performance on detecting and classifying minority classes. Many methods \cite{Li2019GradientHS, Cao2020PrimeSA, Wang2019DataAF, Tripathi2019LearningTG, Cui2019ClassBalancedLB, Tan2020EqualizationLF, Qian2020DRLI} 
have been proposed to tackle this problem from a class imbalance perspective (summarized in Sec \ref{sec_related_work}). We focus on the idea of using loss-reweighting \cite{Cui2019ClassBalancedLB, Tan2020EqualizationLF,Qian2020DRLI} wherein higher weights are assigned to tail classes. Unlike class labels, size (when distinguished as large-to-small) is an ordinal variable making it non-trivial to apply existing solutions for class imbalance to size. Besides, as shown in Figure \ref{fig:vis-size-imb}, small-sized objects are dominant in drone-based aerial datasets and large-sized objects are sparse. Although large-sized objects are the tail, 
they have larger spatial support which can provide richer and more useful features compared to small objects which can make it helpful to detect them. 

On the other hand, learning small-sized objects, although the majority in such datasets, is challenging, even for state-of-the-art detection models \cite{Ren2015FasterRT, Redmon2016YouOL, Lin2017FocalLF}. The increasing use of drone images and the lack of a consistent method for detection of objects of different sizes in such datasets motivates us to solve the severe size imbalance in such aerial datasets. In summary, we address the long-tailed size imbalance issue in drone-based aerial datasets rather than the long-tailed class imbalance issue that is typically addressed in earlier related efforts. 

To this end, we propose a novel architecture-agnostic loss-reweighting strategy which considers the ordinality of the size variable in its design. 
The performance of an object detection model on instances of a given size would have a contribution from object instances of neighboring sizes. For example, given a particular class, a model learned on object instances of area $X$ is more likely to recognize an instance of area $X\pm\delta$ rather than $X\pm k\delta$, where $k$ is a large integer. We hence apply a Gaussian amplification on the size distribution to consider the effect of such neighborhood instances (as detailed in Section \ref{gaussian_amp_section}). 

We subsequently use a clustering approach to assign weights to object instances based on their sizes. 
Finally, inspired by previous balanced loss work which focus on class imbalance \cite{Cui2019ClassBalancedLB}, we reweight the loss based on size clusters to suit our problem. Unlike existing methods for long-tailed class imbalance which assign lower weights to head categories, our method assigns higher weights to the head categories (small-sized objects) ensuring that the model learns better on them. We show that the size-imbalance problem can be addressed using such a loss-based approach without the need for time-consuming architecture engineering.
To summarize, our key contributions are as follows:
\vspace{-1mm}
\begin{itemize}
\itemsep0em
    \item We propose a novel architecture-agnostic loss-reweighting strategy to solve the severe size imbalance issue in drone-based aerial image datasets. 
    We call this \textbf{AR}chitect\textbf{U}re-agnostic \textbf{BA}lanced Loss (\textbf{ARUBA}), which can be applied while training any object detection model. 
    \item To the best of our knowledge, this is the first such loss-based approach to handle size imbalance in this domain. Our key observations around the ordinality of the considered categories and the connection of such ordering to a model's performance may be useful in other settings with ordinal categories (e.g. class labels of a disease with increasing severity levels).
    \item We propose a simple yet effective pipeline based on well-known modules to achieve the objectives using our loss-reweigting strategy. Our extensive experimental results corroborate the usefulness of this pipeline. 
    \item We perform a comprehensive suite of experiments on multiple drone-based aerial image datasets including HRSC2016, DOTA-v1.0, DOTA-v1.5 and VisDrone to validate the effectiveness of our proposed approach. We also provide additional ablation studies and qualitative results to illustrate the usefulness of the proposed method to handle size imbalance in this domain.
\end{itemize}

\section{Related Work}
\label{sec_related_work}
We describe prior work from different related perspectives individually below.

\vspace{0.5em}
\noindent \textbf{Aerial Object Detection.} Compared to the general object detection \cite{liu2020ijcv},  aerial object detection requires special attention because of the additional challenges like high variation in orientation of the objects. 
Specialized methods \cite{yang2019r3det,han2020align,han2021redet} have been designed for detecting oriented bounding boxes in such aerial image datasets. R3Det \cite{yang2019r3det} proposed a feature refinement module for accurate features, thereby improving performance. S$^2$aNet addressed the issue of misalignment between anchor boxes and axis-aligned convolutional features by proposing Feature Alignment Module and Oriented Detection Module. Recently, ReDet \cite{han2021redet} encoded rotation equivariance and rotation invariance by incorporating rotation-equivariant networks. However, all these methods use architecture-based approaches, as mentioned earlier. We instead propose a loss-based approach to address this problem. We, in fact, make use of the abovementioned methods as baselines and show that our loss re-weighting strategy achieves improvement in performance when applied on top of them. 

\vspace{0.5em}
\noindent \textbf{Size Imbalance.} 
There have been fewer efforts that have explicitly addressed size imbalance in object detection as summarized in \cite{oksuz2020imbalance}. These approaches typically depend on using multiple scales of images, feature maps or both to detect objects of different sizes. 
Methods like SSD \cite{liu2016ssd} and Scale-aware Fast-RCNN \cite{li2017scale} make predictions from multiple layers of feature maps and combine them. Feature Pyramid Networks \cite{Lin2017FeaturePN} and its variants aggregates features from multiple layers before performing prediction. Image pyramid-based methods like SNIP \cite{singh2018analysis} and SNIPER \cite{singh2018sniper} use multiple scales of images rather than features for detecting objects of different sizes. \cite{pang2019efficient,li2019scale} combines the advantages of both feature pyramids and image pyramids. The idea behind these methods is to improve the performance by processing at multiple scales. We, on the other hand, exploit the long-tail imbalance of the size distribution by proposing a loss-reweighting strategy for this challenge. 

\vspace{0.5em}
\noindent \textbf{Long-tailed Object Detection.} Existing efforts on long-tailed imbalance generally focus on class imbalance and are divided into three categories: sampling-based, data generation and re-weighting based methods. We describe each of them below.

\vspace{0.5em}
\noindent \textit{Sampling based methods}: Sampling-based approaches rely on data manipulation techniques such as under-sampling and over-sampling. Works such as \cite{Li2019GradientHS,Pang2019LibraRT,Cao2020PrimeSA} utilize sampling-based methods to balance background-foreground and class labels in the dataset. 

\vspace{0.5em}
\noindent \textit{Data generation methods}: These methods generate objects of minority classes synthetically using data generation methods such as Generative Adversarial Networks and data augmentation  \cite{Wang2019DataAF,Tripathi2019LearningTG}. Unlike oversampling, this approach does not repeat data samples and thus reduces over-fitting. However, the performance of these methods is contingent on quality of the samples generated. 

\vspace{0.5em}
\noindent \textit{Re-weighting based methods}: Re-weighting methods formulate the training objective of a model based on the statistics of the class-imbalanced dataset.
\cite{Cui2019ClassBalancedLB} balance the loss based on the effective number of instances per class. \cite{Tan2020EqualizationLF} ignore the discouraging gradients for the rare categories from majority categories. \cite{Qian2020DRLI} alleviate the class-imbalance problem by posing it as a ranking problem. 

These approaches tackle the long-tailed imbalance problem from a class perspective. 
However, we tackle this problem from a size perspective. 

\vspace{0.5em}
\noindent \textbf{Regression Imbalance.}
One of the works closest the present work is DIR \cite{yang2021delving} which focuses on imbalance in continuous targets in general rather than categorical. We focus on continuous targets specific to object detection and propose a framework to mitigate the issue of object size imbalance, which is different from their focus.
\vspace{-0.3em}
\begin{figure*}[!]
    \centering
    \includegraphics[width=0.85\textwidth]{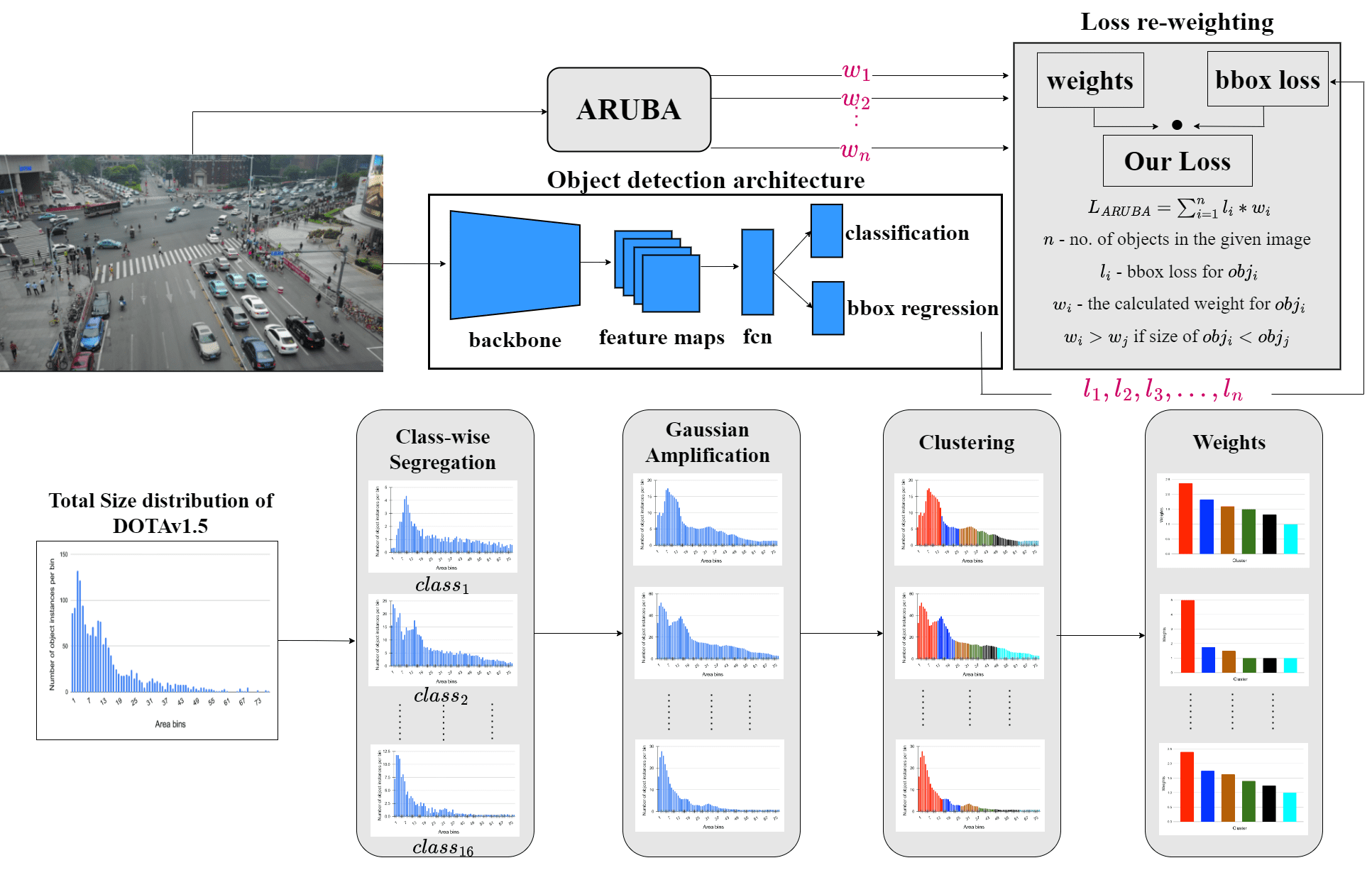}
    \vspace{-2mm}
    \caption{\textbf{Overview of proposed method:} \textbf{(a)} The top figure shows how our method is architecture-agnostic. Independent of the object detection architecture, ARUBA calculates weights for objects based on their sizes. \textbf{(b)} The bottom figure details the ARUBA pipeline comprised of four stages. We use the size distribution of the DOTA\_v1.5 dataset for visualization.}
    \label{fig:archi}
    \vspace{-4mm}
\end{figure*}

\section{Architecture-Agnostic Balanced Loss}
As stated earlier, the proposed \textbf{ARUBA} (\textbf{AR}chitect\textbf{U}re-agnostic \textbf{BA}lanced) loss is designed to address the problem of severe size imbalance in drone-based aerial object datasets. To formulate \textbf{ARUBA}, we begin by discussing the loss re-weighting strategy used in general long-tailed class imbalance methods \cite{lin2017focal, Cui2019ClassBalancedLB, Tan2020EqualizationLF}: 
\begin{equation} \label{cb_loss}
    CB(\textbf{p}, y) = w_y * \mathcal{L}_{cls}(\textbf{p}, y) 
\end{equation}

\noindent where $w_y$ is the weight for a class $y$, 
\textbf{p} is the predicted class probability and $\mathcal{L}_{cls}$ is the classification loss. We instead propose herein a size-balanced loss based on the size of the objects within a class as follows:
\begin{equation} \label{cb_loss}
    \mathcal{L}_{ours} = w_{ys} * \mathcal{L}_{reg}(b', b)
\end{equation} 

\noindent where $w_{ys}$ is the weight for an object of size $s$ belonging to a class $y$; 
$b'$ and $b$ are the predicted and ground-truth bounding boxes, and $\mathcal{L}_{reg}$ is the regression loss. The idea of our re-weighting strategy is to assign higher weights to small-sized objects because they have poor spatial support making it difficult to detect them. 
Note that size being an ordinal variable does not have strict partitions like class categories. Learning to detect objects of a given size does imbue a model with the capability to detect objects of similar sizes (at least partially, as also shown in Table \ref{tab:split_train} and explained in the next paragraph). Also, objects in a dataset may have a large variety of sizes, unlike a fixed number of classes. These differences between categorical and ordinal variables make it non-trivial to directly apply long-tailed class re-weighting strategies to solve size imbalance.
To address these differences, we propose a pipeline of steps, which are simple and well-known, to address size imbalance: class-wise segregation, followed by Gaussian amplification and then clustering. The details of each of these modules are provided in subsequent sections. Figure \ref{fig:archi} outlines our overall pipeline.

 \begin{table}
 \footnotesize
   \begin{center}
    \scalebox{1.1}{
    \begin{tabular}{c|c|c|c}
        \toprule
         \multirow{2}{*}{Trained on} & \multicolumn{3}{c}{\centering Tested on}  \\
         \cline{2-4}
          & Small & Medium & Large \\
         \cline{1-4}
         Small  & \textbf{33.78} & 26.87 & 1.81 \\
         Medium & 7.01 & \textbf{46.01} & 15.26\\
         Large  & 2.56 & 23.53 & \textbf{49.21} \\
         \hline
         All & 17.93 & 29.58 & 38.91 \\  
         \bottomrule
    \end{tabular}
}
\end{center}
   \vspace{-3mm}
  \caption{Performance of baseline on different size bins of HRSC2016 dataset. The train and test sets are divided into three bins - Small, Medium, and Large. ALL bin means we consider the entire train data.}
  \vspace{-5mm}
   \label{tab:split_train}
\end{table}

\vspace{1pt}
\noindent \textit{Effect of neighborhood.} Before describing each of the components in our pipeline, we first show the effect of the ordinality of the size variable through a study. In particular, we discuss the effect of neighborhood on adjacent size bins by experimenting on the HRSC2016 dataset which only has a single class, Ship. We divide both train and test data into three kinds: \textit{small}, \textit{medium} and \textit{large}, based on the object sizes. Table \ref{tab:split_train} summarizes the results (Average Precision values) of a recent aerial object detection method, ReDet \cite{han2021redet}, trained and tested on these categories of objects. We note that the model trained on the \textit{small} train bin (bin containing small-sized objects) performs well on the small test bin, and its performance reduces as we move from small test bin to large. Similarly, the performance of a model trained on the large train bin decreases as we move from large test bin to small. The influence of ordinality of the size categories on model performance is evident in these results. We leverage this neighborhood effect by using a Gaussian amplification process that we describe later in this section. 

\noindent \textbf{Class-wise segregation.} As shown in Figure \ref{fig:archi}, the first stage of our overall pipeline is the class-wise segregation. 
Our empirical studies presented in the supplementary section  suggest that the effect of neighborhood should be considered within a class rather than across-classes.

We hence segregate the size distribution class-wise and deal with size imbalance within each class separately (shown as the first stage in Figure \ref{fig:archi}b).

\vspace{2pt}
\noindent \textbf{Gaussian Amplification.} \label{gaussian_amp_section}
We apply a Gaussian amplification on the size distribution of each class to add the context of the size neighborhood. Similar to Label Distribution Smoothing in DIR \cite{yang2021delving}, we use kernel density estimation to achieve our objective. For each class, we convolve a 1D-Gaussian kernel with the size distribution to obtain a smoothed and amplified distribution. We denote the size distribution of class $c$ as $B^c$ and a discrete Gaussian kernel with window size $w$ as $K_w$. They are defined as follows: 
\vspace{-5mm}
\begin{align}
    B^c &= (b^c_1, b^c_2, .... b^c_m) \\
    K_w &= (k_{-w/2}, ....k_{-1}, k_0, k_{+1},...k_{+w/2})
\end{align}
We design the above discrete Gaussian kernel with certain properties: \textbf{1.} It is an odd-symmetric kernel.  \textbf{2.} The peak of the kernel, $k_0$, is always one. We divide the kernel by its maximum value to achieve this. \textbf{3.} It has two hyperparameters, namely, window size $w$ and variance $\sigma$. $w$ is the width of the Gaussian kernel, i.e. it specifies the number of bins $(b_{-w/2}$ to $b_{w/2})$ that we want to consider from the neighborhood. $\sigma$ specifies the importance that we give to each bin while considering the neighborhood. By increasing $\sigma$, we increase the weight given to each neighboring bin.

\vspace{0.5em}
\noindent We thus define Gaussian Amplification, $GA$, as follows:
\vspace{-3mm}
\begin{equation} \label{gaussian_amp_eq}
    GA(b_k) = \sum_{i=-w/2}^{w/2} k_{i} * b_{k+i}
\end{equation}
\noindent where $b_k$ refers to the size bin in consideration and $k_i$ is the corresponding entry of the Gaussian kernel. For the extremities, the convolution is zero-padded accordingly. Unlike Gaussian smoothing, which can result in reduction of the bin values at times, our procedure always results in amplification by design of the Gaussian filter. We hence call it Gaussian Amplification. For better understanding of its functioning, please go through the  example provided in the supplementary section. 

\vspace{6pt}
\noindent \textbf{Clustering.}
Objects in a dataset usually can be of a wide variety of sizes. One way of categorizing the size distribution is to simply consider each size as a different category. However, this may result in too many size categories. 
We divide the objects into multiple equal-sized bins before applying Gaussian amplification to accommodate the effect of neighborhood. However, owing to the large number of object instances in aerial object datasets \cite{Xia2018DOTAAL,Du2019VisDroneDET2019TV}, this results in a large number of bins. This makes the step of weighting the loss terms of each bin tedious. In order to make the loss reweighting step more feasible, we cluster the instance area distribution (after Gaussian amplification) into a fixed number of clusters, which we can then reweight. In this work, we use a simple $k$-means approach for clustering the distribution into $k$ clusters. Figure \ref{fig:archi}b shows an illustration of the size distribution after clustering the data. As we can observe, objects are grouped as per their sizes. Small-sized objects are clustered together and large-sized objects are clustered together with some intermediate-sized clusters in the middle. We found in our empirical studies that this step provided significant control over the reweighting strategy than merely using equal-sized bins.

\vspace{6pt}
\noindent \textbf{Loss function.} We now describe the actual size-balanced loss itself. As explained earlier, the differences between  ordinal and categorical variables make it non-trivial to apply the existing loss re-weighting strategies used for long-tailed class imbalance in solving size imbalance. We bridge this gap by considering the effect of neighboring sized object instances and forming object clusters based on their sizes. This allows us to obtain weights based on size cluster frequencies. Inspired by \cite{Cui2019ClassBalancedLB}, we define effective number of object instances of a class $y$ belonging to a size cluster $s$ as:
\vspace{-2mm}
\begin{equation}
    E_{ys} = \frac{1-\beta^{GA(n_{ys})}}{1-\beta}
\end{equation}
where $GA$ refers to the \textit{Gaussian Amplification} process as in Eqn \ref{gaussian_amp_eq}, $n_{ys}$ is the number of objects of class $y$ in size cluster $s$, and $\beta \in [0,1)$ is a hyperparameter as defined in \cite{Cui2019ClassBalancedLB}, it controls how fast $E_{ys}$ grows as cluster size $s$ increases.
Depending on the size of the dataset, the value of $GA(n_{ys})$ could be very large, which is indeed the case for aerial datasets. A large value 
causes numerical instability, which is the drawback of \cite{Cui2019ClassBalancedLB}. We mitigate this issue by using an $n^{th}$ root as follows: 
\vspace{-2mm}
\begin{equation}
\label{equation1}
    \tilde{E_{ys}} = \frac{1-\beta^{\sqrt[n]{GA(n_{ys})}}}{1-\beta}
\end{equation}

Using $n^{th}$ root stabilizes the effective numbers without changing the way they ($\tilde{E_{ys}}$) are calculated.

\vspace{0.5em}
\noindent In our overall object detection framework, for an object instance belonging to a class $y$ and size cluster $s$, our loss function $\mathcal{L}_{ours}$ is thus given by:
\vspace{-2mm}
\begin{equation}
\label{loss_eqn}
    \mathcal{L}_{ours} = \mathcal{L}_{C}+ w_{ys}*\mathcal{L}_{R}
\end{equation}
where $\mathcal{L}_{C}$ and $\mathcal{L}_{R}$ represent the  classification and regression loss terms respectively, and $w_{ys}$ is the re-weighting factor based on $\tilde{E_{ys}}$ as given below:
\vspace{-2mm}
 \begin{equation}
\label{loss_eqn}
    w_{ys} = 1 - \frac{1}{\tilde{E_{ys}}}
\end{equation}
Adding the above weights $w_{ys}$ to the object detection loss is the only implementation step required in our framework for any object detection architecture, thus making our approach easy to implement and effective.

\section{Experiments and Results} \label{exp_and_res}
\noindent \textbf{Datasets:} We perform extensive experiments on several popular drone-based aerial image datasets, namely, DOTA-v1.0 \cite{Xia2018DOTAAL}, DOTA-v1.5 \cite{dota1.5}, HRSC2016 \cite{liu2017high} and VisDrone \cite{Du2019VisDroneDET2019TV}. The details of these datasets are shared below.

\vspace{4pt}
\noindent \textit{DOTA-v1.0} \cite{Xia2018DOTAAL}: This is one of the largest aerial image datasets released in 2018 containing 2,806 images and 188,282 object instances. The dataset is divided into train, val and test in $1/2$, $1/6$ and $1/3$ ratios respectively. The aerial images are widespread in 15 different categories namely Plane (PL), Baseball-Diamond (BD), Bridge (BR), Ground-Track-Field (GTF), Small-vehicle (SV), Large-vehicle (LV), Ship (SH), Tennis-Court (TC), Baseball-Court (BC), Storage-Tank (ST), Soccer-Ball-Field (SBF), Roundabout (RA), Harbor (HA), Swimming-Pool (SP) and Helicopter (HC). Small-vehicle class is the majority class while Ground-Track-Field (GTF) is the minority class.

\vspace{4pt}
\noindent \textit{DOTA-v1.5}: \cite{dota1.5}: This was released in 2019 as the succeeding version of the DOTA-v1.0 with an additional category Container-Crane (CC) added to it. Although it is made from the same images as DOTA-v1.0, many additional annotations of very small object instances (less than 10 pixels) are added. It has a total of 403,318 object instances which is more than double the instances present in DOTA-v1.0, making it very distinct. Object detection on DOTA-v1.5 is more challenging than on v1.0 because of the newly added very small instances. Small-vehicle class is the majority class while the newly added Container-Crane class is the minority class.

\vspace{4pt}
\noindent \textit{HRSC2016} \cite{liu2017high}: This is an aerial image dataset that focuses on ship detection. It is comparatively smaller in number, but has variation in object sizes. It has 1061 images divided into 436, 181 and 444 images for training, validation and testing respectively.

\vspace{4pt}
\noindent \textit{VisDrone} \cite{Du2019VisDroneDET2019TV}: This dataset is released as a part of the VisDrone Object detection challenge in 2019. It contains a total of 10209 images divided into 6471, 548 and 3190 for training, validation and testing respectively. Train and validation sets have a combined total of nearly 382,000 object instances that are spread across 10 different categories. 

\noindent \textbf{Evaluation Metrics:} For HRSC2016 and VisDrone datasets, we present the results in the standard COCO format, mAP as the mean of APs@$[.5:.05:.95]$. For DOTA-v1.0 and DOTA-v1.5, following \cite{wang2020centermap, yang2019r3det, han2020align, han2021redet}, we present class wise AP@50 and mAP as the mean of class-wise APs.  

\noindent \textbf{Implementation Details:} Our method is architecture-agnostic and can be applied on top of any architecture proposed for object detection. As we aim to solve the size imbalance issues in drone-based aerial object datasets, for our experiments, we chose two recent state-of-the-art aerial object detection architectures \cite{han2020align, han2021redet} as our baselines. We implement our method using the \texttt{mmdetection} repository. For purposes of fair comparison, we use the same backbone, training schedules, optimizer, learning rate, momentum, weight decay, number of epochs and dataset preparation strategy as used in the baseline methods \cite{han2020align, han2021redet}. For training, we use 4 GTX 1080 Ti GPUs and for inference, we use a single GTX 1080 Ti GPU. 

\begin{table}[htp]
       \centering
        \scalebox{1.00}{%
{
    \begin{tabular}{ c | c }
    \toprule
       Method & mAP \\ 
    \hline
       ReDet ~\cite{han2021redet}  & 70.41 \\
       Ours + ReDet ~\cite{han2021redet}  & 72.42 \\
    \bottomrule
    \end{tabular}%
    }
}
  \vspace{-1mm}
  \caption{Comparisons with the baseline on HRSC2016. 
  }
   \vspace*{-3mm}
   \label{tab:hrsc2016_sota}
\end{table}

\begin{table*}[!htb]
    \begin{center}
    \resizebox{\textwidth}{!}{%
    \begin{tabular}{l|c|ccccccccccccccc|c} \toprule
    Method &backbone &PL &BD &BR &GTF &SV &LV &SH &TC &BC &ST &SBF &RA &HA &SP &HC &mAP \\
    \hline 
    DRN~\cite{pan2020dynamic} & H-104 & 88.91 & 80.22        &43.52 &63.35 & 73.48 & 70.69 & 84.94 & 90.14 &83.85        &84.11 &50.12 &58.41 &67.62 &68.60 &52.50 &70.70 \\
    CenterMap~\cite{wang2020centermap}         & R50-FPN     & 88.88        & 81.24        & {53.15} & 60.65        & 78.62  & 66.55        & 78.10        & 88.83        & 77.80        & 83.61        & 49.36        & {66.19} & {72.10} & \textbf{72.36}  & 58.70        & 71.74        \\
    R$^3$Det~\cite{yang2019r3det}              & R50-FPN    & {88.92} & 77.70        & 46.49        & 71.24        & 72.70        & {77.81} & 79.75        & 90.86        & 81.46        & 83.96        & {57.53} & 59.10        & 65.24        & {70.59} & 51.38  & 71.63        \\
   \hline
    S$^2$aNet~\cite{han2020align}             & R50-FPN     & 89.00        & {80.77}  & 51.77        & 70.91        & 78.52        & 78.01        & {87.19} & 90.86        & 84.99        & 84.64        & 58.45        & 63.60        & 66.39        & 67.90        & 57.92        & {74.06} \\[+0.5ex]
    Ours + S$^2$aNet~\cite{han2020align}             & R50-FPN &89.23&81.07&51.92&70.91&\textbf{78.68}&78.97&87.33&90.89&86.07&\textbf{85.41}&63.20&66.22&66.90&69.82&59.81 & 75.20 \\
    \hline
    ReDet ~\cite{han2021redet}                              & ReR50-ReFPN & 89.34        & {83.03} & {53.83}  & {74.35}  & {77.45} & 83.41  & 87.86  & {90.87} & {87.77} & 85.06 & 62.89        & 62.10        & {75.76}  & 70.58        & 57.93        & {76.15}  \\ 
    Ours + ReDet ~\cite{han2021redet}                             & ReR50-ReFPN & \textbf{89.34}        & \textbf{83.17} & \textbf{54.16}  & \textbf{76.24}  & 78.22 & \textbf{83.42}  & \textbf{87.97}  & \textbf{90.90} & \textbf{87.86} & {85.35} & \textbf{65.39}        & \textbf{66.59}        & \textbf{76.17}  & 70.63        & \textbf{61.69}        & \textbf{77.14}  \\
    \bottomrule
    \end{tabular}%
    }
\end{center}
    \vspace{-4mm}
    \caption{Comparison of our method with the state-of-the-art methods on DOTA-v1.0 OBB Task. The results in bold specify the best result of  each column.}
    \vspace{-2mm}
    \label{tab:dota_sota}
\end{table*}

\begin{table*}[!htb]
   \begin{center}
    \resizebox{\textwidth}{!}{%
    \begin{tabular}{l|cccccccccccccccc|c} \toprule
    Method          &PL& BD & BR & GTF & SV & LV & SH & TC & BC & ST & SBF & RA & HA & SP & HC & CC & mAP \\ \hline
    RetinaNet-O~\cite{lin2017focal}      &71.43&77.64&42.12&64.65&44.53&56.79&73.31&90.84&76.02&59.96&46.95&69.24&59.65&64.52&48.06&0.83&59.16 \\
    FR-O~\cite{ren2017faster}            &71.89&74.47&44.45&59.87&51.28&68.98&79.37&90.78&77.38&67.50&47.75&69.72&61.22&65.28&60.47&1.54&62.00 \\
    Mask R-CNN~\cite{he2017mask}     &76.84&73.51&49.90&57.80&51.31&71.34&79.75&90.46&74.21&66.07&46.21&70.61&63.07&64.46&57.81&9.42&62.67\\
    HTC~\cite{chen2019hybrid}       &77.80&73.67&51.40&63.99&51.54&73.31&80.31&90.48&75.12&67.34&48.51&70.63&64.84&64.48&55.87&5.15&63.40\\[+0.5ex]
    \hline
    ReDet~\cite{han2021redet}    &79.20&82.81&51.92&71.41&\textbf{52.38}&75.73&80.92&90.83&75.81&68.64&49.29&72.03&73.36&\textbf{70.55}&63.33&11.53&66.86 \\
    Ours + ReDet~\cite{han2021redet}         &\textbf{79.85}&\textbf{83.02}&\textbf{52.86}&\textbf{72.73}&52.35&\textbf{75.74}&\textbf{87.18}&\textbf{90.87}&\textbf{81.78}&\textbf{68.68}&\textbf{56.90}&\textbf{73.16}&\textbf{73.41}&70.49&\textbf{65.96}&\textbf{14.34}&\textbf{68.71} \\
    \bottomrule
    \end{tabular}%
    }
\end{center}

   \vspace{-4mm}
   \caption{Comparison of our method with the state-of-the-art methods on DOTA-v1.5 test set OBB Task. 
   }
   \vspace{-4mm}
   \label{tab:dota15_sota}
\end{table*}

\begin{table}[!htb]
   \begin{center}
    \resizebox{0.47\textwidth}{!}{%
    \begin{tabular}{l|c|ccc} \toprule
    Method  & Backbone &AP@50 &AP@75 &mAP \\
    \hline
    RetinaNet~\cite{lin2017focal} & R50 &27.7 &12.7 &13.9\\
    DSHNet~\cite{Yu_2021_WACV} & R50 &30.2 &15.5 &16.1 \\
    \hline
    ReDet~\cite{han2021redet}    &ReR50-ReFPN & 30.86 & 19.50 &18.80 \\ 
    Ours + ReDet~\cite{han2021redet}  & ReR50-ReFPN & \textbf{32.84} & \textbf{21.6} & \textbf{20.32}\\
    \bottomrule
    \end{tabular}%
    }
\end{center}

   \vspace*{-4mm}
   \caption{Comparison of our method with the state-of-the-art methods on VisDrone validation set.}
   \vspace{-4.5mm}
   \label{tab:visdrone_sota}
\end{table}

\noindent \textbf{Results:}

\noindent \textit{HRSC2016.} For our experiments, we use ReResNet50 as the backbone and ReFPN as the neck which were proposed in \cite{han2021redet}. We crop all images in HRSC2016 dataset to $800*512$ and perform horizontal flip augmentation.
Table \ref{tab:hrsc2016_sota} shows our results. Our method obtains a notable performance improvement of \textbf{2.01\%} mAP over the baseline method \cite{han2021redet}.

\vspace{4pt}
\noindent \textit{DOTAv1.0.} For both DOTA-v1.0 and v1.5, the images were cropped to $1024*1024$ and augmented with horizontal flips. Table \ref{tab:dota_sota} summarizes the results of state-of-the-art methods on DOTA-v1.0 OBB task. We apply our method on top of two baselines S$^2$aNet \cite{han2020align} and ReDet \cite{han2021redet}. As observed, our method obtains improvement on top of both baselines, showing the architecture-agnostic nature of our approach. ReDet obtains a performance of $76.15\%$ mAP and our method obtains $77.14\%$ mAP. Our model performs better than all existing state-of-the-art methods. Compared to ReDet, our method improves performance on 12 out of 15 classes, which contain a good mix of both small and large-sized objects. Specifically, on classes `Roundabout (RA)' and `Helicopter (HC)', our method achieves an improvement of $\textbf{4.49\%}$ and $\textbf{3.76\%}$ in AP respectively. We observed that these classes have severe size imbalance which shows the efficacy of our approach.
 
\vspace{4pt}
\noindent \textit{DOTAv1.5.}
Table \ref{tab:dota15_sota} provides a comparison with  state-of-the-art results on DOTA-v1.5 OBB task. ReDet \cite{han2021redet} obtains a performance of $66.86\%$ mAP, while our method obtains $68.71\%$ mAP. Our method achieves a gain of $\textbf{1.85\%}$ mAP. We also obtain improvement for most of the classes on this dataset. Specifically, for the class `Basketball Court', which has severe imbalance in object sizes, we obtain an improvement of $\textbf{5.97\%}$ in AP. Despite the fact that DOTA-v1.5 contains a lot of newly added small instances when compared to DOTA-v1.0, our methods achieves better results on DOTA-v1.5, which supports our claim that our method improves performance on small objects. 

\vspace{4pt}
\noindent \textit{VisDrone.} 
We use the same image cropping and augmentation techniques as used for the DOTA datasets. A performance comparison between the state-of-the-art models and our model is given in Table \ref{tab:visdrone_sota}. As the evaluation server for this challenge dataset is closed, we present our model's performance on the validation set, and do the same for the baseline models for fairness of comparison. Our model achieves a performance gain of $\textbf{1.5\%}$ mAP over the baseline. 

\begin{table}[htp]
\footnotesize
   \begin{center}
    \scalebox{1}{
    \begin{tabular}{c|c|c|c|c}
        \toprule
         \multirow{2}{*}{Trained on} & \multirow{2}{*}{Method} & \multicolumn{3}{c}{\centering Tested on}  \\
         \cline{3-5}
           & & Small & Medium & Large \\
         \cline{1-5}
           \multirow{2}{*}{HRSC2016} & ReDet & 17.93 & 29.58 & 38.91 \\
         & Ours + ReDet  & \textbf{20.79} & \textbf{29.97} & 38.01 \\
         \cline{1-5}
         \multirow{2}{*}{DOTA-v1.0} & ReDet & 09.74 & 23.48 & 52.44 \\
         & Ours + ReDet  & \textbf{11.81} & 23.34 & 52.24 \\
         \cline{1-5}
         \multirow{2}{*}{DOTA-v1.5} & ReDet & 8.32 & 24.85 & 43.56 \\
         & Ours + ReDet & \textbf{10.65} & 24.76 & 43.52 \\
         \cline{1-5}
         \multirow{2}{*}{DOTA-v1.0} & S$^2$aNet & 10.64 & 24.93 & 47.43 \\
         & Ours + S$^2$aNet  & \textbf{12.48} & \textbf{25.57} & \textbf{47.85} \\
         \bottomrule
    \end{tabular}
}
\end{center}
  \vspace{-3mm}
  \caption{Comparison between the performance of our model and the baseline model on small, medium and large sized objects.}
    \vspace{-7mm}
   \label{tab:sml_val}
\end{table}

\vspace{4pt}
\noindent \textit{Results on small, medium and large objects.}
Table \ref{tab:sml_val} shows a comparison in the performance of the baselines \cite{han2020align, han2021redet} and our model on 
different sized objects. For these experiments, we use the test set of HRSC2016 and the validation set of DOTA-v1.0 and DOTA-v1.5.  Note that the ground truth annotations of the test set for DOTA dataset are not publicly available, hence, we use the validation set. We follow the same evaluation metrics as mentioned in Section \ref{exp_and_res}. Our method when applied on top of ReDet \cite{han2021redet}, achieves an improvement of $\textbf{2.86\%}$, $\textbf{2.07\%}$ and $\textbf{2.33\%}$ mAP on the small sized objects of HRSC2016, DOTA-v1.0 and DOTA-v1.5 datasets respectively (first three rows of Table \ref{tab:sml_val}). We also provide the performance gain of our method applied on top of a different architecture \cite{han2020align} (last row of Table \ref{tab:sml_val}). On all the datasets, our model maintains the performance on medium and large sized objects as well which shows the efficacy of our approach.

\begin{table}[htp]
\footnotesize
      \centering
{
    \begin{tabular}{l|c|c|c|c}
    \toprule
    Method &FPN &AP@50 &AP@75 &mAP \\ 
    \hline
    S$^2$aNet~\cite{han2020align} & \xmark & 56.27 & 23.72 & 27.85 \\
    Ours + S$^2$aNet~\cite{han2020align} & \xmark & 57.68 & 25.22 & 28.78 \\
    S$^2$aNet~\cite{han2020align} & \cmark & 74.06 & 36.88 & 40.28\\
    Ours + S$^2$aNet~\cite{han2020align} & \cmark & 75.20 & 38.76 & 41.04 \\
    \bottomrule
    \end{tabular}%
    }
  
  \caption{Performance of our model with and without FPN. 
  }
  \vspace{-3mm}
   \label{tab:fpn}
\end{table}

\section{Discussion and Analysis}
\subsection{Ablation studies}
To clearly evaluate the effectiveness of our proposed approach, we perform ablation studies on the DOTA-v1.0 dataset using two baselines S$^2$aNet \cite{han2020align} and ReDet \cite{han2021redet}. We use the ResNet50-FPN and ReResNet50-ReFPN backbones for experiments on the baseline methods \cite{han2020align}  and \cite{han2021redet} respectively. Table \ref{tab:ablation_ga} shows the results, and indicates consistent improvement over the baseline methods.

\begin{table}[htp]
\small
      \centering
{
    \begin{tabular}{l|c|c|c|c}
    \toprule
    Method &GA &AP@50 &AP@75 &mAP \\ 
    \hline
   S$^2$aNet~\cite{han2020align}  &\xmark &74.06 &36.88 &40.28 \\
    Ours + S$^2$aNet~\cite{han2020align}  &\xmark&74.32 &37.12 &40.35 \\
    Ours + S$^2$aNet~\cite{han2020align} &\cmark &\textbf{75.20} &\textbf{38.76} &\textbf{41.04} \\
    \hline
    ReDet ~\cite{han2021redet} &\xmark &76.15 &50.75 &47.05 \\
    Ours + ReDet ~\cite{han2021redet} &\xmark &76.47 &51.15 &47.42 \\
    Ours + ReDet ~\cite{han2021redet} &\cmark &\textbf{77.14} &\textbf{52.93} &\textbf{48.13} \\
    \bottomrule
    \end{tabular}%
    }
   \caption{Performance comparison of our model with and without Gaussian Amplification. GA refers to Gaussian Amplification.     
   }
   \vspace{-4mm}
   \label{tab:ablation_ga}
\end{table}

\noindent \textbf{Effect of Gaussian Amplification: } Table \ref{tab:ablation_ga} shows the performance of our model with and without employing the Gaussian amplification step. As observed, when Gaussian amplification is not applied, results show minimal improvement on both baselines \cite{han2020align, han2021redet} in contrast to a healthy improvement when it is applied. This shows the importance of considering the effect of neighborhood when dealing with ordinal variables like size of objects. 

\begin{figure}[ht]
\centering
\begin{subfigure}{.22\textwidth}
    \centering
    \includegraphics[width=0.9\linewidth]{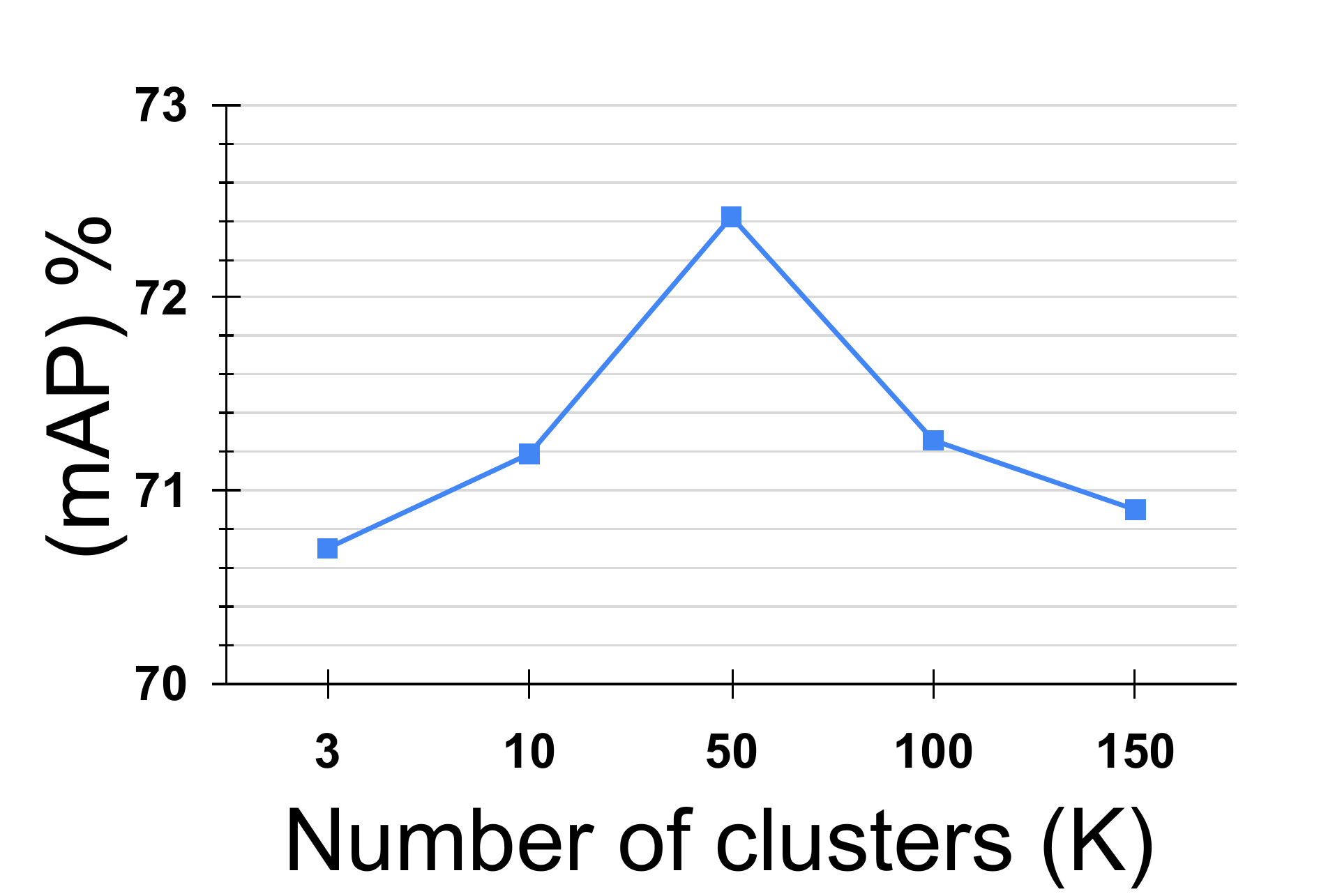}  
    \caption{Number of clusters ($K$)}
    \label{fig:ablation-k}
\end{subfigure}
\begin{subfigure}{.22\textwidth}
  \centering
  \includegraphics[width=0.9\linewidth]{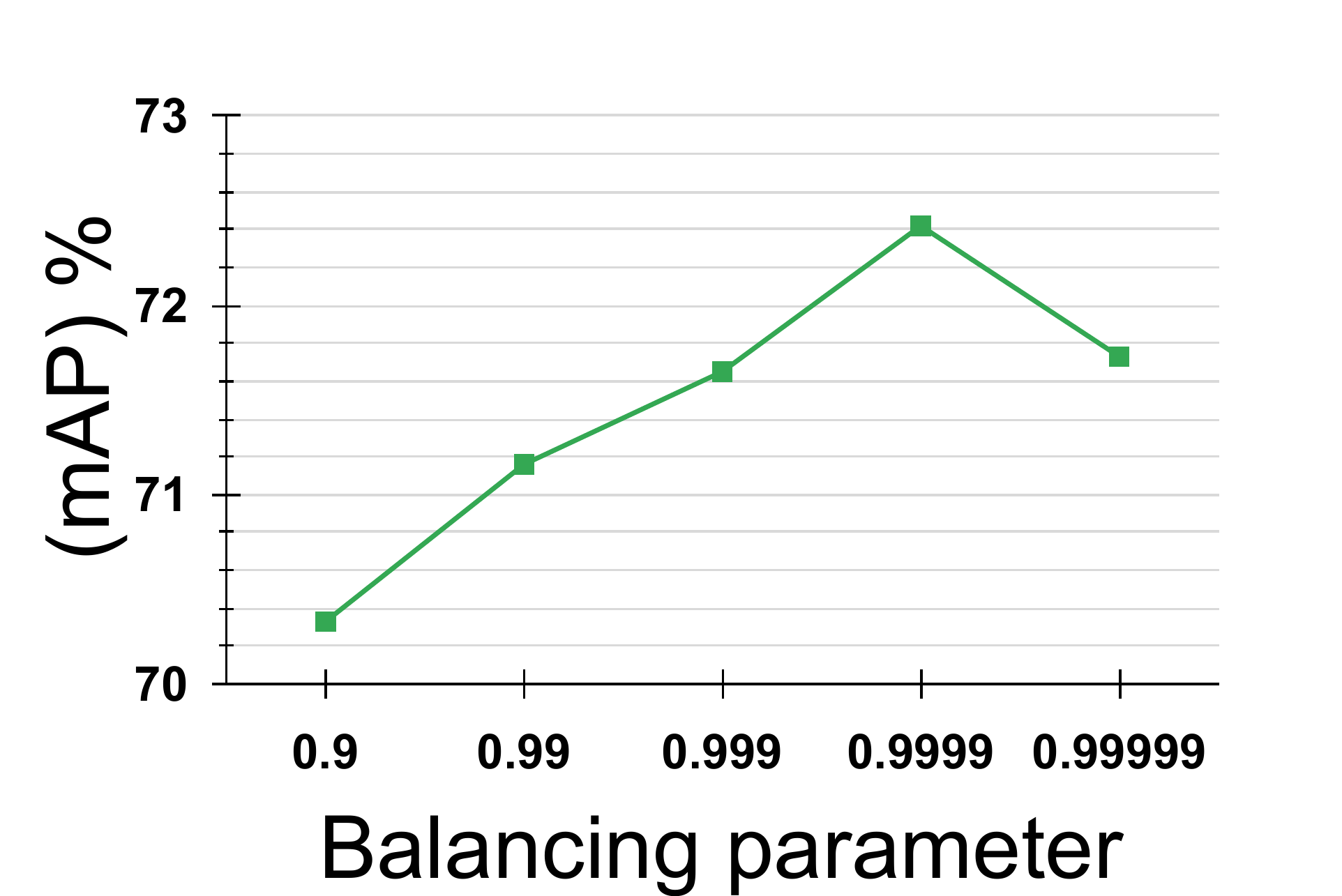} 
  \caption{Balancing parameter ($\beta$)}
  \label{fig:ablation-beta}
\end{subfigure}
\begin{subfigure}{.22\textwidth}
  \centering
   \includegraphics[width=0.9\linewidth]{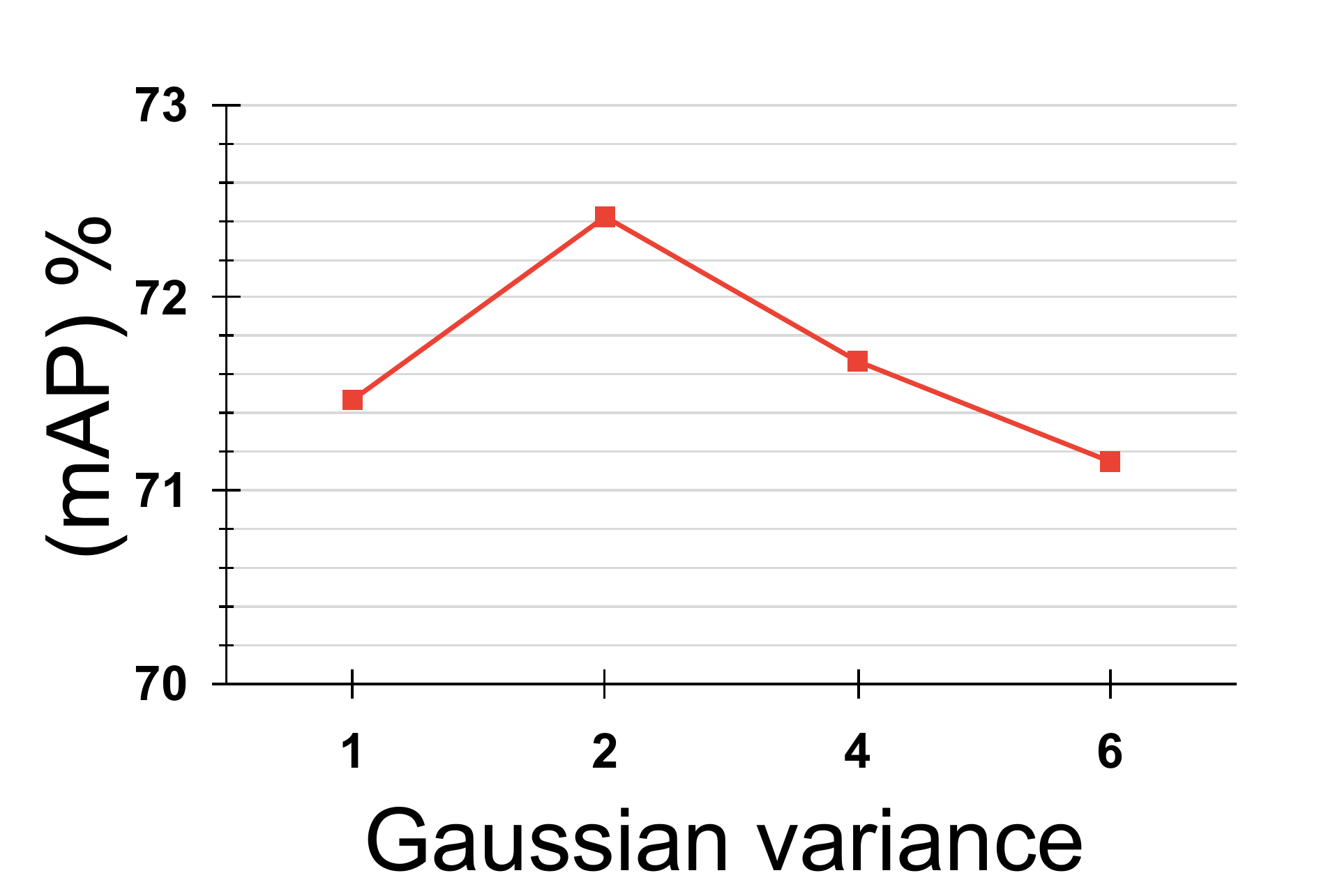}  
    \caption{Gaussian variance $\sigma$}
    \label{fig:ablation-sigma}
\end{subfigure}
\begin{subfigure}{.22\textwidth}
  \centering
  \includegraphics[width=0.9\linewidth]{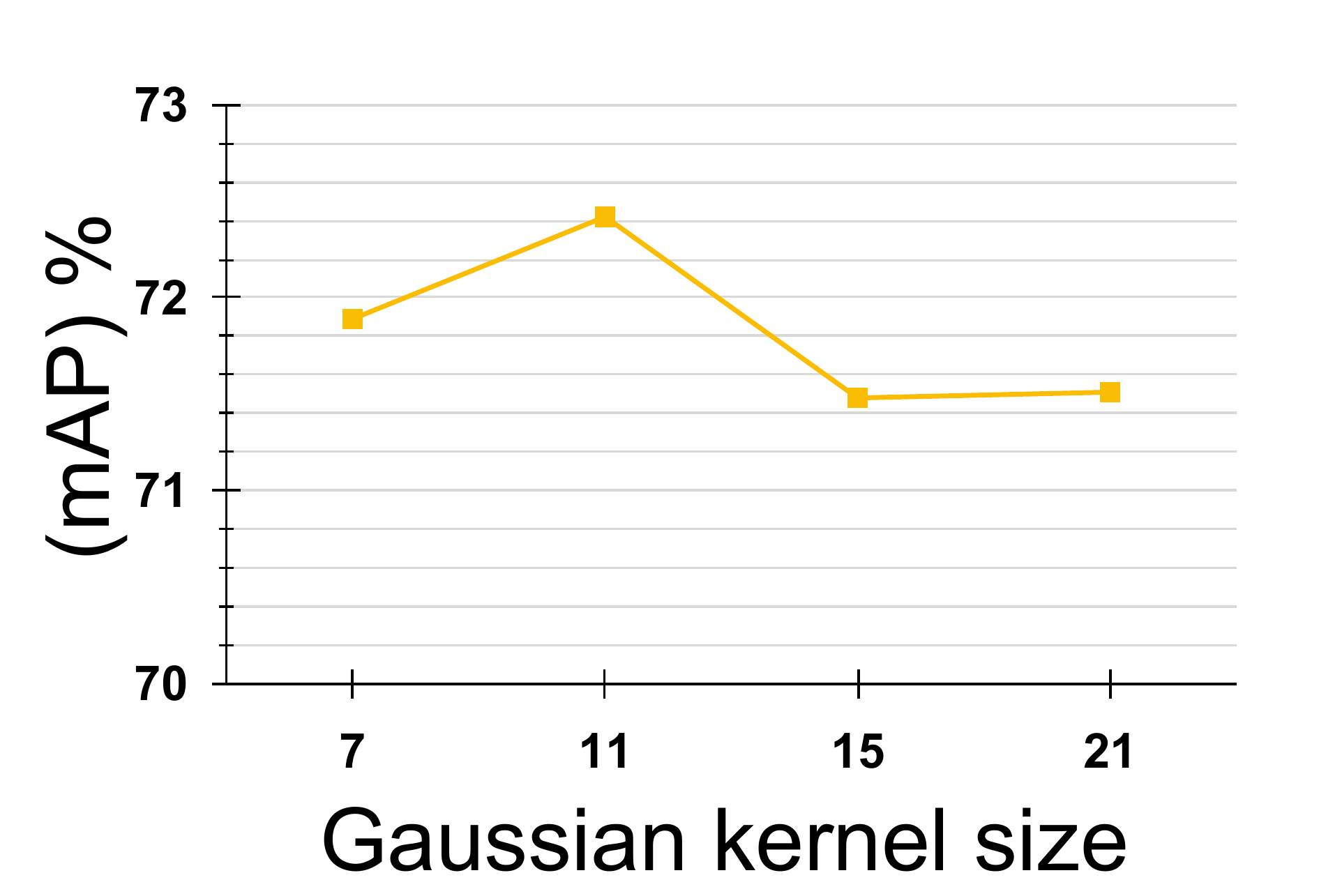}  
    \caption{Gaussian width/kernel size}
    \label{fig:ablation-kernel}
\end{subfigure}

\caption{Effect of various hyperparameters on model performance}
\vspace{-4mm}
\label{fig:hyperparameter-analysis}
\end{figure}

\vspace{-2mm}
\subsection{Hyperparameter Sensitivity Analysis}

\noindent \textbf{Analysis on $k$:} $k$ is the hyperparameter associated with $k$-Means clustering. It determines the number of clusters. We experimented with different $k$ values on HRSC2016 and $k=50$ gives the best results as evident from Figure \ref{fig:ablation-k}. We noticed this to be consistent with other datasets too.

\vspace{3pt}
\noindent \textbf{Analysis on $\beta$:} $\beta$ is the hyper-parameter used in the calculation of effective weights. To determine the best value of $\beta$, we vary $\beta$ in the range $[0.9, 0.99,0.999,0.9999, 0.99999]$. Figure \ref{fig:ablation-beta} shows the performance on HRSC2016 dataset. We observed that by increasing the value of $\beta$, starting from $0.9$, the performance increases until $\beta=0.9999$ which achieves the best result. Lower values of $\beta$, eg. 0.5, 0.6, and 0.7  make the effective weights uniform. This issues is also cited in \cite{Cui2019ClassBalancedLB}. Upon experimentation on smaller $\beta$s, we observed that the results were close to the baseline. 

\vspace{0.25em}
\noindent \textbf{Analysis on $\sigma$:} $\sigma$ specifies the level of importance to be given to neighboring bins when applying Gaussian amplification to a given bin. It can also be regarded as an amplification factor. Figure \ref{fig:ablation-sigma} shows the performance of our model as we change the $\sigma$ value. We obtained the best results when $\sigma$ is set to $2$. 

\vspace{0.25em}
\noindent \textbf{Analysis on $w$:} This hyperparameter specifies the number of neighboring bins to consider while applying Gaussian amplification to a given bin. We experimented on HRSC2016 dataset by varying $w$, refer Figure \ref{fig:ablation-kernel}. We found that a width of 11 gives the best results.

\vspace{0.5em}
\noindent We note that these hyperparameter values performed well across all the four datasets considered, without the need to fine tune the model separately on different datasets. 


\subsection{Further Analysis}
\vspace{-4pt}
Feature Pyramid Network \cite{Lin2017FeaturePN} is one of the primary methods developed to solve the problem of high variation in sizes of instances in object detection datasets. Many methods \cite{pang2019efficient,li2019scale} have been proposed based on the idea of FPN. We perform experiments on DOTA\_v1.0 dataset to show that our approach improves detection performance when applied on top of FPN. Table \ref{tab:fpn} summarizes the results of these experiments. As observed, when FPN is not applied, our model improves mAP from 27.85 to 28.78. Also, when FPN is used, our model achieves the best performance by improving mAP from 40.28 to \textbf{41.04}.  FPN enhances the model's capability to process at multiple scales while our method addresses the severe long-tail in the size imbalance. Hence, best results are obtained when our method is applied on top of such architecture-engineered methods.

\begin{figure}[t]
    \centering
    \includegraphics[width=\linewidth]{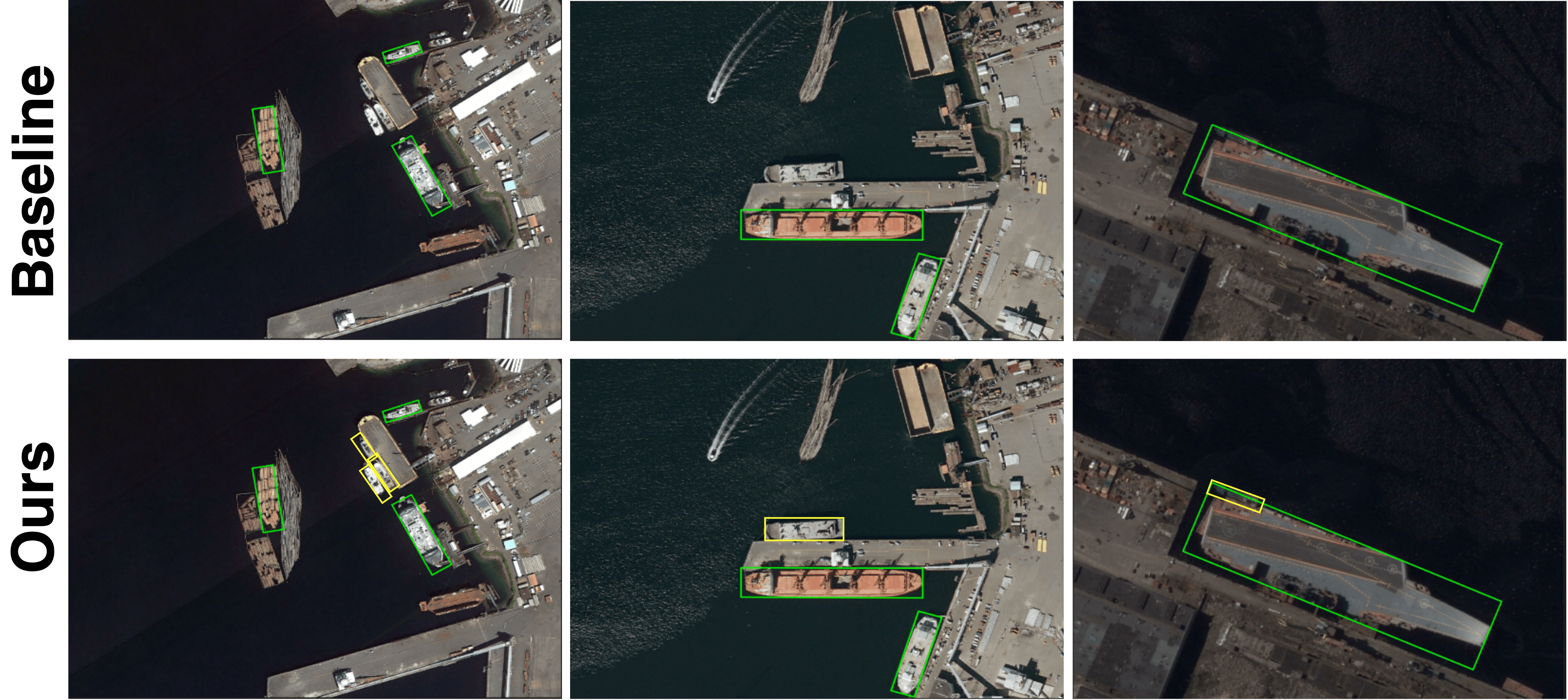}
    \caption{Predictions on images from HRSC2016 dataset \cite{liu2017high} - ReDet  \cite{han2021redet} vs Ours. \textbf{Top:} The baseline method fails to detect small sized objects. \textbf{Bottom:} Ours is able to recognize additional small objects. Yellow boxes indicate objects additionally detected. }
    \label{demo}
    \vspace{-3mm}
\end{figure}

\vspace{-3mm}
\section{Conclusions and Future Work}
\vspace{-4pt}
In this work, we presented a framework to alleviate imbalance in the object size distribution. We proposed a novel simple-to-implement architecture-agnostic loss re-weighting method for drone-based aerial object detection. We dealt with the ordinality of size by taking into consideration the effect of neighborhood instances on prediction and by clustering the object instances based on their size. We showed the need to increase the contribution of small objects despite them belonging to the head of the long-tail size distribution. 
We showed that our method improves performance on popular datasets like HRSC2016, DOTAv1.0 DOTAv1.5 and VisDrone. In future, we plan to extend this work by mitigating the imbalance of class and size together.  

\noindent \textbf{Acknowledgements.} We are grateful to the Ministry of Electronics and Information Technology and Ministry of Education, Govt of India, as well as IIT-Hyderabad through its MoE-DRDO fellowship program for their support of this project. We also thank the anonymous reviewers and Area Chairs for their valuable feedback in improving the presentation of this paper.


{\small
\bibliographystyle{ieee_fullname}
\bibliography{egbib}
}

\end{document}